\documentclass[final,3p,times,twocolumn]{elsarticle}

\usepackage{amssymb}
\usepackage{amsmath}
\usepackage{amsfonts}
\usepackage{array}
\usepackage{graphicx}
\usepackage{caption}
\usepackage{subcaption}
\usepackage{xcolor}
\usepackage[normalem]{ulem}
\usepackage{booktabs}
\usepackage{hyperref}

\newcommand{\modelname}{${\rm KeGCN}_{R}$}
\newcommand{\backboneModel}{KeGCN}

\newcommand\torevise[1]{\textcolor{black}{#1}}

\useunder{\uline}{\ul}{}

\journal{Knowledge Based Systems}

\begin{document}

\begin{frontmatter}



\title{Corporate Fraud Detection in Rich-yet-Noisy Financial Graph}


\author[nju-ai,nju-skl]{Shiqi Wang} 
\author[nju-ai,nju-skl]{Zhibo Zhang}
\author[nju-sem]{Libing Fang}
\author[nju-ai,nju-skl]{Cam-Tu Nguyen}
\author[nju-cs,nju-skl]{Wenzhong Li}

\affiliation[nju-ai]{organization={School of Artificial Intelligence, Nanjing University},
            addressline={}, 
            city={Nanjing},
            postcode={}, 
            state={Jiangsu},
            country={China}}
\affiliation[nju-cs]{organization={School of Computer Science and Technology, Nanjing University},
            addressline={}, 
            city={Nanjing},
            postcode={}, 
            state={Jiangsu},
            country={China}}
\affiliation[nju-sem]{organization={School of Management and Engineering, Nanjing University},
            addressline={}, 
            city={Nanjing},
            postcode={}, 
            state={Jiangsu},
            country={China}}
\affiliation[nju-skl]{organization={State Key Laboratory for Novel Software Technology, Nanjing University},
            addressline={}, 
            city={Nanjing},
            postcode={}, 
            state={Jiangsu},
            country={China}}
\begin{abstract}
Corporate fraud detection aims to automatically recognize companies that conduct wrongful activities such as fraudulent financial statements or illegal insider trading. \torevise{Previous learning-based methods fail to effectively integrate rich interactions in the company network.} To close this gap, we collect 18-year financial records in China to form three graph datasets with fraud labels. 
We analyze the characteristics of the financial graphs, highlighting two pronounced issues: (1) \textit{information overload}: the dominance of (noisy) non-company nodes over company nodes hinders the message-passing process in Graph Convolution Networks (GCN); and (2) \textit{hidden fraud}: there exists a large percentage of possible undetected violations in the collected data. The hidden fraud problem will introduce noisy labels in the training dataset and compromise fraud detection results. 
To handle such challenges, we propose a novel graph-based method, namely, Knowledge-enhanced GCN with Robust Two-stage Learning (\modelname), which leverages Knowledge Graph Embeddings to mitigate the information overload and effectively learns rich representations. 
The proposed model adopts a two-stage learning method to enhance robustness against hidden frauds.
Extensive experimental results not only confirm the importance of interactions but also show the superiority of {\modelname} over a number of strong baselines in terms of fraud detection effectiveness and robustness.

\end{abstract}


\end{frontmatter}



\section{Introduction}
Corporate fraud refers to illegal schemes by listed companies in the stock market, aiming at financial gains through different means such as fraudulent financial statements and illegal insider trading. This kind of fraud bears systematic risks, which can potentially lead to financial crises at the macro level \cite{li2010case}. Unfortunately, the rapid growth of young capital markets has given rise to an increasing number of fraudulent cases in recent years, putting pressure on regulators and auditors. Since the traditional human supervision solution is no longer efficient, it is desirable to build an autonomous system to assist regulators in this essential task. 

\begin{figure}
    \centering
    \includegraphics[width=0.95\linewidth]{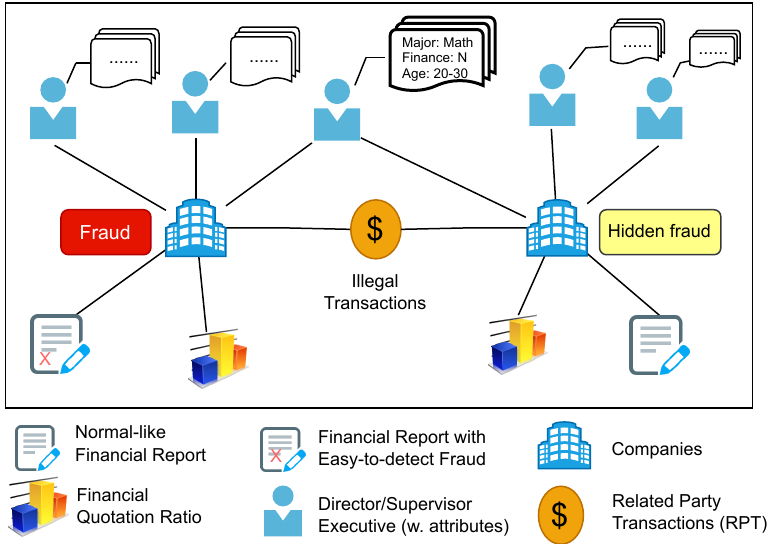}
    \caption{Relations (e.g. illegal transactions) are essential for corporate fraud detection. When a violation goes undetected in the historical record, it is referred to as a hidden fraud case. Such cases become label noises hindering the effectiveness of corporate fraud detection.}
    \label{fig:1}
\end{figure}

Previous studies on machine learning for corporate fraud detection focus mostly on traditional machine learning (ML) methods such as linear regression \cite{dechow2011predicting,dalnial2014detecting,hajek2017mining}, random forest \cite{song2014application,hajek2017mining}, and BP neural network \cite{song2014application,hajek2017mining}. These machine-learning models are built to classify annual financial statements as fraudulent or not, based on expert-chosen feature sets. Unfortunately, the rich interactions in the company network have not been effectively integrated for corporate fraud detection.  Financial experts, on the other hand, have recognized the influence of  ``Directors/Supervisors/Executives (DSE)''  and ``Related Party Transactions (RPT)'' on corporate fraud (see Figure \ref{fig:1}). DSE refers to the members of the director board of the company. Being the decision-making body in a company, the director board is certainly the agent behind most corporate frauds \cite{Hatice2004}. Connection via DSE also helps companies lower the coordination cost for illegal activities, thus significantly increasing the likelihood of committing fraud \cite{KHANNA2015}. RPT refers to deals or arrangements between two companies that are joined by a previous business association or share common interests. RPTs, particularly those that go unchecked, carry the risk of financial fraud by various means such as illegal profit transmission \cite{Jian2010, CLAESSENS20061}.

In order to better understand the role of rich relations between companies in fraud detection, we first collect 18-year financial records of A-share listed companies with fraud labels from the Chinese stock market and design our financial knowledge graph. The companies are categorized by their board markets into ``Main-Board Market'' (MBM), ``Growth Enterprise Market'' (GEM), and ``Small and Medium Enterprise Board Market'' (SME). Additionally, DSE and RPT instances are collected and used to connect company instances to form 3 financial graphs for MBM, GEM, and SME, respectively. Thorough data analysis (see section \ref{sec:data analyse}) is then conducted on the collected datasets, revealing two main challenges that hinder effective data integration in the financial graph.\\ 

\noindent\textbf{Challenge I (Information Overload).} Our objective is to incorporate relational data on a financial graph to help corporate fraud detection. The common approach is to make use of Graph Convolution Networks \cite{kipf2016semi,liu2020towards,yun2022graph,yu2022multiplex}, which have shown great success in dealing with relational data. Unfortunately, directly applying GCN \cite{kipf2016semi} on financial graphs is difficult. This is because the number of DSE and RPT nodes in our graph (referred to as support nodes) dominates the number of company nodes (referred to as target nodes) by around 20 times, hindering the message-passing process between companies. In addition, many of the support nodes lack attributes, making the problem more serious. We refer to this as the information overload problem, which is caused by the excessive number of (noisy) support nodes. Unfortunately, existing graph-based methods for other fraud detection tasks  \cite{hu2019cash,ji2022detecting,xu2021towards,li2021live,xu2021towards} are not of help since they do not face this problem. More specifically, these studies target personal or transaction fraud detection, where the number of these target nodes (persons, transactions) is from 4 to 100 LARGER than the support nodes. The closest one to ours is \cite{bi2022company} which exploits the hierarchical graph structure to carefully distill information from the support nodes into the target nodes, mitigating the information overload issue. This method, however, is not general enough to be applied to our task due to the strong graph structure assumption.\\


\noindent\textbf{Challenge II (Hidden Fraud).} In the real world, regulators often discover a company's fraud much later than the time it occurred.  According to our collected data in the 18 recent years, only 30\% of frauds were found in the same year as the violation actually happens (see Section \ref{sec:data analyse}), which implies a large percentage of ``hidden fraud'' that cannot be detected timely.  Those hidden frauds will result in noisy labels in the training dataset. It has been proven that noisy labels can cause serious bias in the GCN models and harm the classification results severely \cite{qian2022robust,dai2021nrgnn}. However, the hidden fraud problem was rarely addressed in the literature and a robust GCN model is desirable to deal with this issue.



In this paper, we propose a novel Knowledge-enhanced GCN with a Robust two-stage Learning framework, or \modelname{}, that systematically targets the above challenges. Specifically, for the first challenge, we leverage knowledge graph embeddings (KGE) methods \cite{bordes2013translating,sun2018rotate,DGL-KE} to distill useful information from support nodes to company nodes by exploiting (sample) pairwise relations.  For example, KGE could help bring a company embedding closer to an RPT embedding if there should be a link (a pairwise relation) between them, or further away if there should not be any link. These company embeddings are then considered as another view besides financial attributes to be used for learning with GCN on company sub-graphs, which contain only company instances and are constructed with meta-paths \cite{sun2011pathsim,wang2019heterogeneous}. Since KGE exploits contrastive learning with sample links to predict pairwise semantic relations instead of aggregating information from the neighborhood like GCN, it is less likely to suffer from the information overload issue. In addition, because a company subgraph contains only company instances, the message-passing process between companies becomes more effective. Note that our solution is highly adaptable to other tasks since we do not have any assumption on the graph structure, and the selection of meta-paths can be adapted to fit other requirements.


For the second challenge, we design a novel robust two-stage learning method based on learning Bayes optimal distribution \cite{yang2022estimating}. The main idea is to estimate a Bayes transition matrix, which encodes the probabilities of Bayes optimal labels (predicted labels of optimal classifiers) to noisy labels for a company instance. With the estimated transition matrix, one can perform loss correction to learn a label noise robust model. Unlike previous works \cite{yang2022estimating,zhu2021second,cheng2020learning,cheng2020learning2}, however, hidden fraud has two distinct characteristics: 1) It has an asymmetric noise structure, which is based on the fact that regulators prefer precision in declaring frauds with the cost of letting some fraud cases go undetected (hidden frauds); 2) It is instance and neighborhood dependence (see Figure \ref{fig:1}). By considering such issues, our method is better designed to mitigate hidden fraud. All in all, our work has the following contributions:



\begin{itemize}
    \item We construct new financial graphs from 18-year records of Chinese A-share companies datasets, larger than previous datasets on corporate fraud detection. Our dataset incorporates rich information such as ``Related Party Transaction'' (RPT) and ``Director/Supervisors/Executives'' (DSE) from the company network for corporate fraud detection. To facilitate further research in this field, we have made all datasets and associated code publicly available.\footnote{https://github.com/wangskyGit/KeHGN-R}.

    \item We investigate two unexplored problems of hidden frauds and information overload in corporate fraud detection. Detailed analysis reveals that these two problems exist in real datasets and have a strong side effect to the goal of fraud detection. 
    
    \item We propose a novel method, namely Knowledge-Enhanced GCN with Robust Two-stage Learning ({\modelname}), that is specifically designed to handle the distinct characteristics of our problem. Our experiments show that {\modelname} is more effective in comparison to several contemporary graph-based models and label-noise robust methods.
\end{itemize}
    
    
    
\section{Related study}
\subsection{Graph neural networks.}
GNN has received growing interest in recent years thanks to its ability to handle relational data. Graph convolutional neural networks (GCNNs) adapt convolutional neural networks to graph data, effectively integrating node features and graph topology \cite{bronstein2017geometric, li2018deeper}. \cite{kipf2016semi} propose GCN, a simple version of GCNNs that applies an efficient approximation for fast convolution layers. GraphSAGE \cite{hamilton2017inductive} can be viewed as a stochastic generalization of graph convolutions, and it is especially useful for massive, dynamic graphs that contain rich feature information. DAGNN \cite{craja2020deep} is a deeper GNN model that decouples the transformation and aggregation of message passing. Graph Relearn Network (GRN) \cite{grn} introduces a two-phase approach to enhance prediction stability. In its initial pre-predict phase, followed by a relearn phase, the network progressively improves predictions for nodes that show instability. GsCP  \cite{gscp} framework employs strategic graph augmentation techniques to generate multiple views of node embeddings. This method simplifies the process by avoiding complex similarity calculations while achieving superior performance compared to other contrastive learning approaches in capturing structural features that generalize well. Recently, methods to handle heterogeneous information on graphs have been proposed based on, e.g., meta-path \cite{wang2019heterogeneous,yun2019graph,yun2022graph}, relation embedding \cite{schlichtkrull2018modeling,shang2019end,yu2021knowledge}, edge modeling \cite{zhang2019heterogeneous,hu2020heterogeneous}. Unfortunately, these methods are not designed to tackle the distinct challenges in our problem.\\

This work focuses on the challenge of information overload that can occur when using Graph Neural Networks (GNNs) on graphs with a specific characteristic. This type of graph has a large number of noisy support nodes, which outnumber the target nodes. This is commonly seen in real-world situations, such as the examples provided in two related works, Bi et al. \cite{bi2022company}, who attempt to predict company financial risk using shareholders as support nodes, and Liu et al. \cite{liu2021intention}, who use user behavior sequences of actions such as browsing, clicking to detect fraudulent transactions. Our work differs from these studies in that we are the first to characterize and empirically demonstrate the ineffectiveness of GNNs on graphs with information overload. Moreover, we propose a simple-yet-effective method to handle this issue by distilling information from support nodes to target nodes and then sampling subgraphs of target nodes for GNN. Our approach can be applied to a wide range of graphs with information overload, while previous methods \cite{bi2022company, liu2021intention} have more restrictive graph structure assumptions. Specifically, Bi et al. \cite{bi2022company} assume a hierarchical graph structure whereas Liu et al. \cite{liu2021intention} suppose that support nodes (user behaviors) form a sequence structure.

\subsection{Corporate fraud detection.}
Corporate fraud detection has been an important topic in finance since the early days of capital markets. Linear methods such as the MScore model \cite{beneish1999detection} and FScore model \cite{dechow2011predicting} are among the most widely adopted methods early on. Recent studies \cite{dechow2011predicting,dalnial2014detecting,hajek2017mining,craja2020deep,song2014application} apply traditional ML models with expert-chosen financial attributes and achieve notable success. However, there has been limited work that studies the effectiveness of exploiting relational indicators such as DSE \cite{KHANNA2015} and RPT \cite{Jian2010}. Only Mao et al. \cite{mao2022financial} have considered RPT relations for corporate fraud detection. Specifically, the authors represented companies and RPT relations in a knowledge graph, similar to our approach. The authors, however, manually extracted features from the knowledge graph and then applied traditional machine-learning methods. In particular, neither knowledge graph embeddings nor graph neural networks have been used in their study \cite{mao2022financial}. In recent years, some studies also considered using GNN methods in financial fraud detection \cite{motie2024financial}, however, most of them simply leverage modern GNN methods and didn't consider the hidden-fraud or information-overload problem which we focus on.  

Also relevant to our work are cash-out fraud detection \cite{hu2019cash,ji2022detecting,mienye2024deep}, consumer loan fraud detection \cite{xu2021towards} and online fraud detection \cite{liu2022user,li2021live,xu2021towards}, where graph-based methods have been successfully exploited. The most essential difference lies in the fact that they do not face the information overload problem as we do. In addition, the hidden fraud problem has not been addressed in these studies.


\subsection{Learning with label noise.} 
It is desirable to develop machine learning models to be robust against label noise in training data, which is very common in practice. Based on the assumption of the noise type, current methods can be divided into models for random noise \cite{natarajan2013learning,manwani2013noise},  class-dependent noise \cite{liu2015classification,patrini2017making,yao2020dual}, or instance-dependent noise \cite{zhu2021second,cheng2020learning,yang2022estimating,hao2022model}. Few studies \cite{cheng2020learning,qian2022robust} in recent years have tried to address noise associated with relational data on graphs. Unfortunately, these solutions can not fully handle our hidden fraud issue, of which noise depends not only on class (asymmetric noise structure) but also on the instance and its neighborhood. 


In this study, we define and characterize the problem of hidden fraud, a comparably unexplored class of label noise on graphs. Here, hidden fraud is a type of label noise that is asymmetric, instance-and-neighbor dependence noise. This type of label noise is commonly found in many fraud detection problems as it is difficult for labelers to detect all the violations in a timely manner. Despite the growing interest in label noise on graph data \cite{xu2021towards,qian2022robust,cheng2020learning}, we are the first to provide a real-world graph dataset with label noise.

\section{Preliminary}\label{sec:preliminary}
\torevise{Given all the important relations and properties, it is crucial to consider how we can better use these various types of data and build the financial knowledge graph}. This section introduces our knowledge graphs and the concept of meta-path company subgraphs. These are preliminary concepts that will be used to describe our proposed model, \modelname.



\subsection{Financial Knowledge Graph}

A knowledge graph (KG) \cite{rossi2021knowledge} is a multi-relational graph that represents knowledge as a set of entities and the relationships between them. The entities in a KG are represented by vertices ($\mathcal{V}$), and the relationships between them are represented by edges ($\mathcal{E}$). Each edge is labeled with a relation type from a predefined set $\mathcal{R}$. Formally, a KG can be represented by a set of triples $\{(h_i,r_i,t_i)\}$, each of which is a tuple of three elements: the head entity $h_i\in \mathcal{V}$, the relation $r_i\in \mathcal{R}$, and the tail entity $t_i\in\mathcal{V}$. \\

\begin{figure}
    \centering
    \includegraphics[width=0.95\columnwidth]{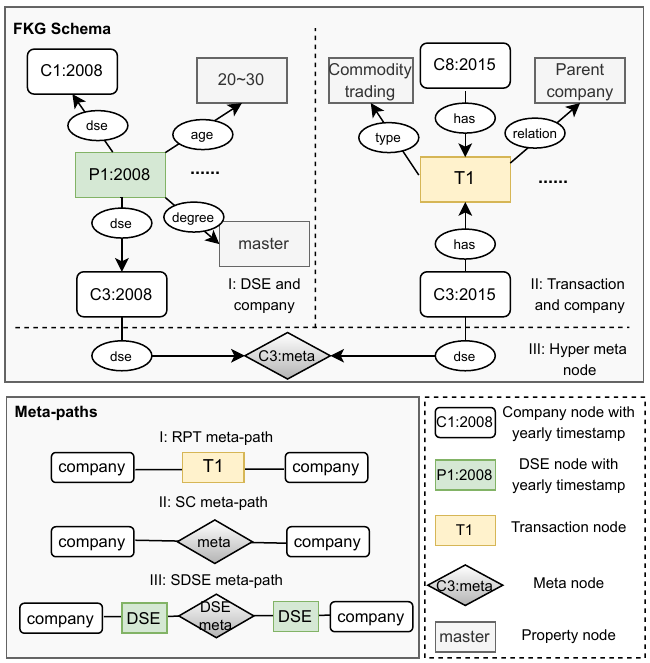}
    \caption{The Schema of FKG and our predefined meta-paths.}
    \label{fig:kg}
\end{figure}

\noindent\textbf{Definition 1.} Financial Knowledge Graph (FKG) \textit{is a knowledge graph obtained from the company network using the following conventions (see FKG Schema in Figure \ref{fig:kg}): 1) The entity set $\mathcal{V}$ contains company, DSE, and RPT instances, as well as attributes of DSE such as education, positions and RPT such as transaction type;  2) The relation set $\mathcal{R}$ contains relations defined for companies such as (company, has, transaction), as well as those defined for RPT or DSE such as (transaction T1, type, commodity trading).}\\

It is important to note that in a FKG, a company or a DSE corresponds to multiple entities, each of which represents the company or the company director in a specific year. This is because the task of corporate fraud detection is performed on annual company reports, whereas companies (or directors (DSE)) in different years have different attributes. To avoid information loss, we connect company or DSE entities of different years to a common $meta$ or \textit{DSE-meta} entity node.


\subsection{Meta-path based Company Subgraphs}
Meta-paths are commonly used in GNNs to help in learning heterogeneous information \cite{wang2019heterogeneous,yun2019graph,yun2022graph}. Here, we adopt this popular idea and use company meta-paths to represent different possible relations between companies.\\

\noindent\textbf{Definition 2.} A Meta-Path\cite{sun2011pathsim} \textit{$A_1\stackrel{R_1}{\longrightarrow} A_2 \dots \stackrel{R_l}{\longrightarrow} A_{l+1}$ is defined as a composite relation from $R_1$ to $R_l$.} A Company Meta-path \textit{is a meta-path such that the starting node ($A_1$) and the ending node ($A_{l+1}$) are both company entities.} \\

This paper considers three types of meta-paths: RPT meta-path, SC meta-path and  SDSE meta-path. Different meta-paths reveal different relations existed among companies. RPT meta-path, which corresponds to the pattern \textit{company-transaction-company}, captures the relation that two companies ``had a related party transaction.'' SC meta-path connects instances of ``the same company in different years.'' Finally, the SDSE meta-path conveys the semantic that two companies ``share the same DSE.'' 

Given a company meta-path, one can construct a new  subgraph, where an edge between two companies is formed if there is a company meta-path leading from one to the other. However, in FKG, there might be multiple paths of the same meta-path type connecting a pair of companies. For this reason, we define a Multi-path Weight Matrix, which can be considered as the edge weights of the subgraph. \\


\noindent\textbf{Definition 3.} A Multi-path Weight Matrix \textit{is constructed for each company meta-path $\rho_k$. It is a matrix $\mathbf{W}^{mp}_{k}$ of size $N^c\times N^c$ where $N^c$ is the number of companies in FKG. Each element $(i,j)$ of $\mathbf{W}^{mp}_{k}$ is defined as the number of possible paths between company nodes $i,j$ in FKG that match $\rho_k$.} \\

\noindent\textbf{Definition 4.}  A Meta-path based Company Subgraph \textit{$\mathcal{G}_k=\langle\mathcal{V}^c, \mathcal{E}_k, \mathbf{W}^{mp}_k\rangle$ is a graph with only company nodes connected by a company meta-path $\rho_k$, where $\mathcal{V}^c\subset\mathcal{V}$ denotes the company nodes in FKG,  $\mathcal{E}_k$ denotes edges following the pattern in $\rho_k$, $\mathbf{W}^{mp}_{k}$ is the multi-path weight matrix of $\rho_k$, which plays the role of edge weights in $\mathcal{G}_k$.} \\

Here, we consider only company meta-paths, which can avoid introducing an excessive number of support nodes during message-passing in GNNs. It is noteworthy that we do not ignore the support nodes during training, the information will be exploited during the knowledge embedding learning.

\section{Data Collection and Analysis}\label{sec:data analyse}
This section describes our data collection process, and provides more information on the inherent challenges of our problem.


\subsection{Data Collection and Statistics}
 Since previous datasets are neither large enough nor contain company relations \cite{hajek2017mining,craja2020deep,song2014application}, we collect our data by combining relevant information from several databases on CSMAR\footnote{https://cn.gtadata.com/}. More specifically, we first extract all the annual financial records of listed companies on two stock exchange markets (Shanghai and Shenzhen) during the period from 2003 to 2020. Data after 2020 might contain much noise due to the  We then combine the financial statements and the stock market quotation data to extract 429 financial attributes for each company in a specific year. The financial attribute set includes all kinds of financial ratios, the annual value, the volatility of monthly value for turnover rates, and the return of stock prices, to name a few. Finally, we map RPT, and DSE records from the RPT and DSE databases to the corresponding company instance, where each RPT (DSE) record contains 12 (10) attributes. 
 
 \begin{table}[t]
\centering
\caption{Statistics for three collected datasets in the Chinese stock market from the year 2003 to 2020.}
\resizebox{\columnwidth}{!}{%
\scalebox{1}{
\begin{tabular}{@{}cccc@{}}
\toprule
\multicolumn{1}{l}{} & \textbf{MBM} & \textbf{SME} & \textbf{GEM} \\ \midrule
\#Company nodes      & 23341       & 12651        & 9422         \\
\#DSE and RPT nodes  & 658702       & 220848       & 147061         \\
\# fraud activity & 6351        & 3351         & 1640         \\
\#fraud: \#non-fraud & 1: 6.87     & 1: 5.25      & 1: 8.35      \\
FKG: \#entities      & 790374      & 295461       & 220060       \\
FKG: \#relations     & 41          & 41           & 41           \\
FKG: \#edges     & 5580916          & 2311465           & 1651936           \\
Year range           & 2003-2020   & 2003-2020    & 2009-2020    \\ \bottomrule
\end{tabular}%
}
}

\label{tab:dataset-statistics}
\end{table}

Following financial studies \cite{lee2004information,zhu2019comparative}, we categorize companies based on their board markets for research and comparisons. We consider three main markets which together account for the majority of listed companies, including the ``Main Board Market'' (MBM), ``Small and Medium Enterprise'' (SME) and ``Growth Enterprise Board Market'' (GEM). We then build one FKG for each market according to the procedure in Section \ref{sec:preliminary}. The resulting FKGs with statistics are shown in Table \ref{tab:dataset-statistics}, from which several insights can be drawn. Firstly, MBM is the biggest board market with more than 0.66 million entities and more than 20 thousand company instances. In reality, it is also the most important market in China, hence requiring higher standards in terms of capital stock size, and business terms, to name a few. Secondly, GEM is the smallest one with only 9422 company instances. In practice, GEM also has more lenient listing requirements than the other two. Note also that the difference in the year range of GEM  is due to the fact that it was officially established in 2009. And finally, the number of supporting nodes (DSE and RPT) is 28 times larger than the company nodes in MBM, 18 times larger in SME, and 16 times larger in GEM. Additionally,
94.71\% of DSE instances and 100\% RPT instances contain
at least one missing attribute. The excessive number of noisy support nodes causes the problem of information overload as stated in the introduction.


\subsection{Hidden Fraud Analysis}
\begin{figure}[t]
    \centering
        \begin{subfigure}[b]{0.75\columnwidth}
    \includegraphics[width=\textwidth]{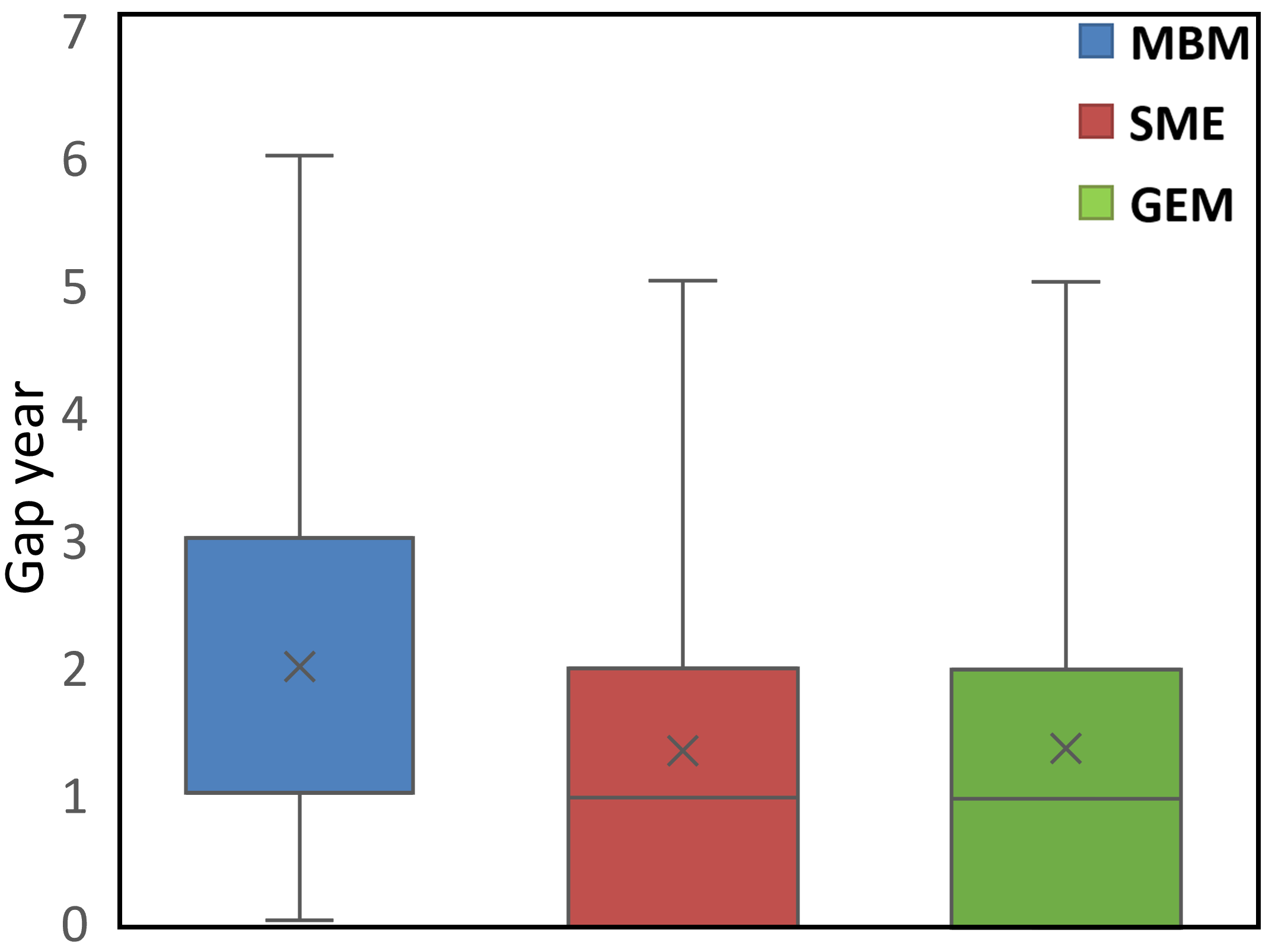}
    \caption{The year gap distributions in MBM, SME, and GEM;}
    \label{fig:mbfraud}
\end{subfigure}
\\
    \begin{subfigure}[b]{0.75\columnwidth}
        \includegraphics[width=\textwidth]{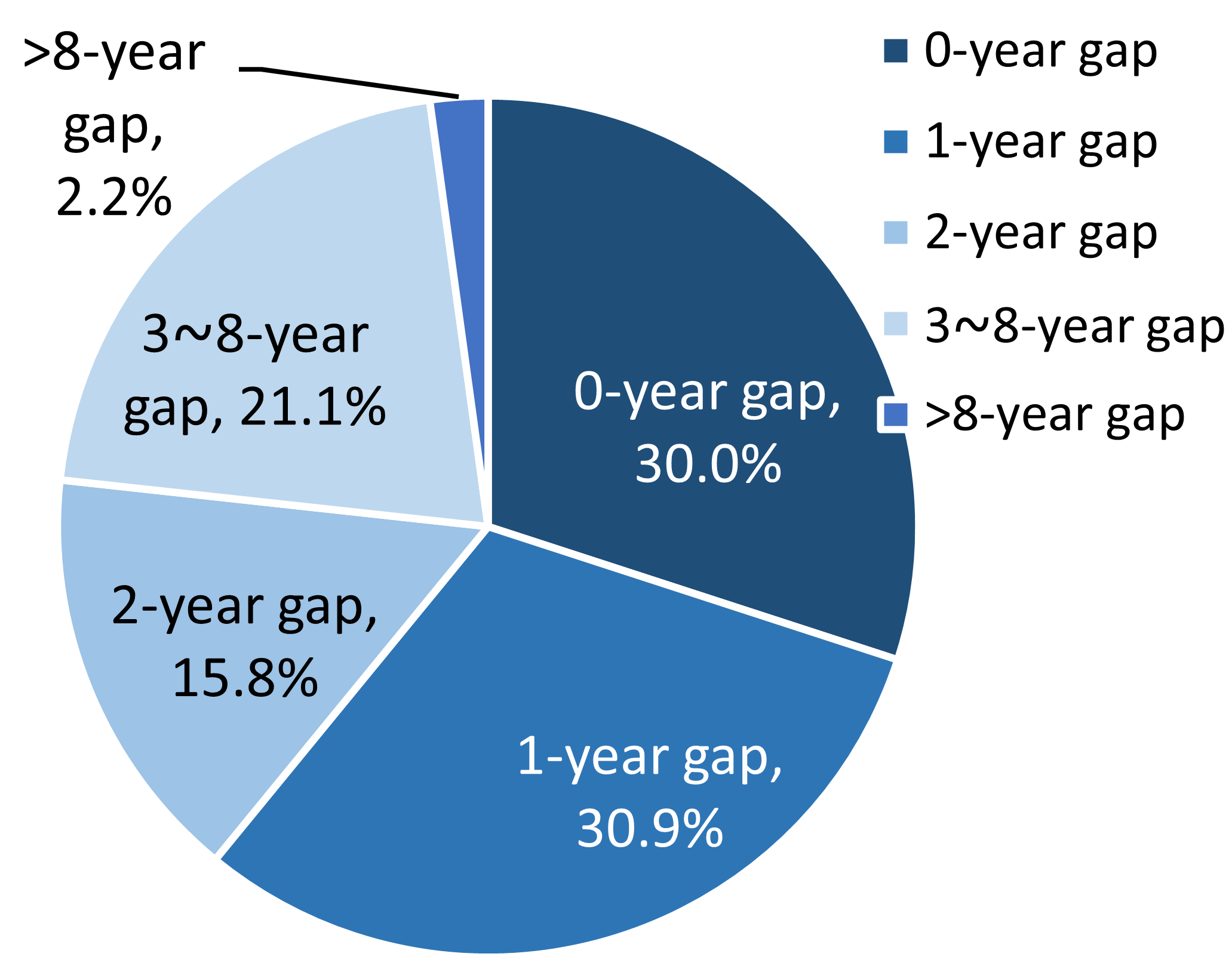}
        \caption{The percentages of year gap ranges calculated from all three datasets.}
   
    \end{subfigure}
    \caption{ Statistic of datasets }
    \label{fig:fraud}
\end{figure}

To study the impacts of hidden frauds, we collect two date values associated with each company fraud case: 1) the violation date which is the time when the fraud activity actually happens; and 2) the declared date which is when regulators declare the fraud and the detailed information to the public. We then define the year gap as the difference in the year of these two values. The distributions of year gaps in three datasets are plotted in Figure \ref{fig:fraud} (a) and the percentages of different gap ranges are shown in Figure \ref{fig:fraud} (b).

Figure \ref{fig:fraud} shows that the problem of hidden fraud is serious, with only 30\% of all the cases (in all datasets) can be detected in the same year. The average year gap in MBM is around 2 years, bigger than that of the other two datasets (see Figure \ref{fig:fraud}), showing that the problem of hidden fraud is most serious in the MBM dataset. In addition, 50\% of violations have gone undetected for at least 2 years in MBM and at least 1 year in SME and GEM.  Note that outliers with much longer delays are not shown in this figure, for which the delay in detection could be 16 years in MBM, 11 years in SME and 10 years in GEM.


In light of such observations, it is inevitable that there are hidden frauds among those labeled as non-fraudulent ones in the collected data. A general machine learning model trained on such kind of noisy labeled data is likely to fail to detect fraud in a timely manner, possibly leading to unwanted consequences. Unfortunately, studying this problem is non-trivial due to the lack of an evaluation benchmark. Previous studies on noisy labeled data often follow two ways to create benchmark datasets: 1) We can start with a clean dataset and then actively introduce noises for research. Usually, researchers need to have some assumptions about the noise distribution or the noise ratio. This method cannot apply to our case since we neither have a clean dataset nor know enough about the hidden fraud distribution for a sufficiently good simulation. In addition, actively corrupting fraud data points into hidden fraud ones can cause the class imbalance problem to be more serious, making it hard to control the actual effect of label-noise robust methods; 2) We can ask humans to check and verify wrongly labeled cases in a previously labeled dataset like the Clothing1M dataset \cite{xiao2015learning}. This is, however, also unrealistic in our case since examining corporate fraud requires much higher expertise than other applications like fashion/animal image annotation.

To handle the aforementioned issue, we further study the year gap distribution in Figure \ref{fig:fraud} (b), where we observe that only 2.2\% fraud cases are undetected after 8 years. This leads us to an assumption that we can reasonably trust a case that has been labeled as non-fraud for at least 8 years and consider it as a clean non-fraud data point. As a result, the procedure to construct our hidden fraud benchmark is as follows: (1) The training sets for experiments are sampled from our datasets without any filtering. This way, we do not change the characteristics of the real-world data, and the objective is to build a model that is robust to the existence of the unknown noise in training; (2) For testing, however, we filter non-fraud cases within 8 recent years, leaving us with considerably clean test sets for evaluation. As far as we know, these are also the first real-world graph-based datasets with label noise.

\section{The Base Model: Knowledge-enhanced GCN} \label{sec:base-method}
 The problem of corporate fraud detection is formalized as a binary classification that aims to assign labels $y$ to company nodes in FKG with $y=1$ being fraudulent and $0$ otherwise. More specifically, we are given  $\mathcal{G}^{kg}=\left\langle\mathcal{V},\mathcal{R},\mathcal{E}\right\rangle$, where the company nodes in $\mathcal{V}^c \subset \mathcal{V}$ are associated with financial attributes $\mathbf{X}^{att}\in R^{N^c\times d^{att}}$, and the objective is to generate the fraud possibility vector $\mathbf{y}_v \in \mathcal{R}^{1\times2}$ for each company node $ v \in \mathcal{V}^c$. The elements of $\mathbf{y}_v$ represent for the possibility of the node being normal and fraud respectfully.

\begin{figure*}[t]
    \centering
    \includegraphics[width=0.95\linewidth]{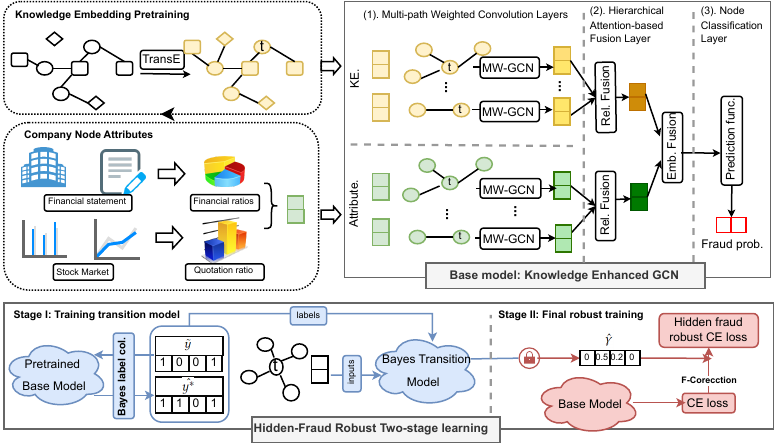}
    \caption{The model (\modelname{}) architecture (best seen in colors). The base model \backboneModel{} is shown in the right-upper corner. It contains two branches with either knowledge embedding or financial attributes as initial node embedding. For simplicity and efficiency, we choose to use TransE for knowledge embeddings. Embeddings from different meta-path subgraphs and different initial embedding types are then fused with the attention mechanism. Steps for hidden-fraud robust two-stage learning are shown at the bottom of the image, where $\tilde{y}$ stands for the noisy labels in the datasets, $\hat{y}^*_i$ denote the estimated Bayes optimal labels by an independent training base model. After getting $\hat{y}^*_i$, it together with $\tilde{y}$ are then leveraged to train the Bayes transition model. During stage II, the trained  Bayes transition model is fixed and predicts hidden fraud rate $\hat{\gamma}$ for F-correction.}
    \label{fig:model}
\end{figure*}

Our base model architecture is shown in Figure \ref{fig:model}, which consists of four phases: 1) \textbf{Knowledge embedding learning}: this is the pretraining phase,  where KGE methods are used to learn company embeddings $\mathbf{X}^{ke}$ from FKG; 2) \textbf{Multi-path weighted convolution layers}: company meta-paths are exploited to draw meta-path based company subgraphs $\{\mathcal{G}_k|k\in[1,N^{mp}]\}$. For each $\mathcal{G}_k$, there are two ways to initialize node representations, i.e. either $\mathbf{X}^{att}$ or $\mathbf{X}^{ke}$. Subsequently, node representations are then learned on different subgraphs by using Multi-Path Weight Graph Convolutional Networks (MW-GCN); 3) \textbf{Hierarchical attention-based fusion}: Two kinds of attention are exploited to combine node representations from subgraphs, which take into account the importance of different meta-paths and different initialization strategies ($\mathbf{X}^{att}$ or $\mathbf{X}^{ke}$);  4) \textbf{Node classification layer}: The fraud probability $\mathbf{y}_v \in \mathcal{R}^{1\times2}$ (for $v$) is inferred by a single neural network layer.  

\subsection{Knowledge Embedding Pretraining}

KGE learns to map entities and relations into continuous vector space and facilitates the task of predicting unknown triples in the knowledge graph. In other words, it aims to learn entity and relation embeddings from triples in our FKG $\mathcal{E}=\{(h_i,r_i,t_i)\}$. Unlike GNN, which propagates and aggregates information from neighbors, KGE methods focus on the interactions of entities via pairwise relations. In general, many KGE methods can be applied in our model, such as TransE \cite{bordes2013translating}, DistMult \cite{yang2015embedding}, ComplEx \cite{trouillon2016complex}, RotaTE \cite{sun2018rotate}. For simplicity and efficiency, we choose to use TransE in this paper.

The main idea of TransE is to map entities and relations to the same vector space, say $R^{d^{ke}}$, so that we can connect entities $h$ and $t$ via the relation $r$ using $h+r\approx t$ where $(h,r,t)$ is an observed triple (or fact) in our KG. For example, we can obtain embeddings for transaction ``T1'', the relation ``type'', and the attribute entity ``Community Trading''  so that ``T1'' + ``type'' $\approx$ ``Community Trading''. Formally, TransE learns a scoring function $f$ as follows:
$$f(h,r,t)=-||h+r-t||_{1/2}$$
where $||_{1/2}$ is either $L_1$ or $L_2$ norm. The scoring function is larger if $(h,r,t)$ is more likely to be a fact, i.e., $h$ is connected to $t$ via $r$ in KG. Contrastive learning \cite{hadsell2006dimensionality} is used to learn embeddings for all the entities and relations by enforcing the scores of true triples to be higher than those of negative (distorted) triples.

After the pretraining phase, we extract embeddings $\mathbf{X}^{ke}\in R^{N^c\times d^{ke}}$ of all the company instances. The relation constraints in FKG allow TransE to distill useful semantic information from embeddings of connected nodes and relations to company embeddings.  

\subsection{Multi-path Weighted Convolution Layers}
This phase aims to learn representations from company subgraphs $\mathcal{G}_k=\langle\mathcal{V}^c, \mathbf{W}_k^{mp}, \mathbf{X}^{it}\rangle$, each obtained from a meta-path $\rho_k$ and an initialization strategy ($\mathbf{X}^{att}$ or $\mathbf{X}^{ke}$). For simplicity, we drop $k$, $it$ and show how higher representations are learned from a single graph.

We develop MW-GCN, a new variant of GCN \cite{kipf2016semi} that takes into account the multi-path weights. Let the representations from $l$-th layer be $\mathbf{H}^l\in R^{N^c\times d_l}$ ($\mathbf{H}^0=\mathbf{X}$), where the $v$-th column vector is the representation of $v\in\mathcal{V}^c$ at $l$-th layer and $d_l$ is the hidden size of the $l$-th layer. For learning the representations of $l+1$ layer, row normalization is first conducted on $\mathbf{W}^{mp}$:
$$\mathbf{\hat{W}}^{mp}(i,j)=\frac{\mathbf{W}^{mp}(i,j)}{\sum_{l=1}^n \mathbf{W}^{mp}(i,l)}$$
Second, for the target node $v$, the $v$-th row of $\mathbf{\hat{W}}^{mp}$ is used to obtain an aggregated representation from the $l$-th layer representations $\mathbf{h}_u^l\in R^{d_l}$ from the neighbors of $v$: 
$$a_v^{l+1}=\sum_{u\in\mathcal{N}_{\mathcal{G}}(v)}\mathbf{\hat{W}}^{mp}(v,u)\mathbf{h}_u^l$$
which is then combined with the hidden representation of $v$:
$$\mathbf{h}_v^{l+1}=\sigma[\textbf{W}_l(a_v^{l+1}+\mathbf{h}_v^l)]$$
where $\sigma$ is the nonlinear function, $\mathbf{h}_v^{l}\in R^{d_l}$ and $\mathbf{h}_v^{l+1}\in R^{d_{l+1}}$ are the representations of $v$ at layer $l$ and $l+1$, $\mathcal{N}_{\mathcal{G}}(v)$ is the set of neighbours of $v$ in a subgraph $\mathcal{G}$ and $\textbf{W}_l$ is the learning parameter. 

Compared to GCN, WM-GCN introduces $\mathbf{W}^{mp}$ to take into account the importance of meta-paths for aggregating (meta-)neighbors. This resembles the idea of attention in GAT \cite{velickovic2017graph} and HAN \cite{wang2019heterogeneous}. However, the multi-path weights in WM-GCN are simply drawn from the domain knowledge (FKG graph) instead of learning, making WM-GCN comparably more efficient.

\subsection{Hierarchical Attention-based Fusion Layer}

The input for this phase is a set of representations $\{\mathbf{H}^{L}_{it,k}\in R^{N^c, d_L}\}$ from the  output layers of WM-GCN trained on $N^{sg}=2\times N^{mp}$ subgraphs; where $k \in [1,...,N^{mp}]$ and $it\in \{ke, att\}$. For simplicity, all WM-GCN are set to have the same number of layers and the same output dimension $d_L$. The objective of this fusion step is to pool the input representations to obtain $Z\in R^{N^c, d_{Z}}$ for prediction.\\

\noindent\textbf{Relation Attention SubLayer.} We pool the representations from different subgraphs with the same initialization strategy $it$, which is dropped from the notation for simplicity. Inspired from HAN \cite{wang2019heterogeneous}, for each $H^L_k$, we leverage a readout function to get the graph embedding and the relation attention value as:
$$\mathbf{h}_{\mathcal{G}_k}=\frac{1}{N^c}\sum^{N^c}_{i=1} \mathbf{h}^L_{k,i}$$
$$a^{rel}_k=\frac{\exp(\sigma(\mathbf{W}^{rel} \mathbf{h}_{\mathcal{G}_k}+b^{rel}))}{\sum_{k\in N^{mp}}\exp(\sigma(\mathbf{W}^{rel} \mathbf{h}_{\mathcal{G}_k}+b^{rel}))}$$
where $\sigma$ is the activation function, $\mathbf{W}^{rel}\in R^{1\times d_{L}}$, $b^{rel}\in\mathbb{R}$ are shared for all meta-path based subgraphs. We then obtain the relation-aware representation for $v$ as follows:
 
 $$\mathbf{z}^{rel}_v=\sum_{k\in N^{mp}}a^{rel}_k \mathbf{h}_{k,v}^{L}$$
 \\
 
\noindent\textbf{Embedding Attention SubLayer.} This layer calculates $Z$ from $Z^{rel}_{ke}$ and $Z^{rel}_{att}$, the relation-aware representations with $ke$ and $att$ initialization strategies. Using a similar mechanism as the relation attention layer, we learn the attention weights, which are then used to calculate a weighted sum of $Z^{rel}_{ke}$ and $Z^{rel}_{att}$ to obtain $Z\in R^{N^c\times d_Z}$.


\subsection{Node Classification Layer}
Given the fusion representation $Z=[\mathbf{z}_1,...,\mathbf{z}_{N^c}]$, where $\mathbf{z}_v$ is the representation of $v\in\mathcal{V}^c$, a fully connected layer is used to get the probability of node $v$ to be fraudulent:
$$\mathbf{\hat{y}}_v=\text{sigmoid}(W^{pred} \mathbf{z}_v+b^{pred})$$
where $\langle W^{pred}, b^{pred}\rangle$ are the layer parameters.

\section{Hidden-Fraud Robust Two-stage Learning}
As we described before, hidden fraud issues are crucial in corporate fraud detection. In this section, we describe how we handle this issue \textbf{during the training phase} in a model-agnostic way.

Before introducing our method, we first formalize our problem into the framework of learning Bayes optimal distribution as follows:.\\

\noindent\textbf{Hidden Fraud Robust Learning Problem.} Considering a company node $v$ with attributes $x_v$ and $\tilde{y_v}$ being its (noisy) label in the dataset, we denote the true (clean) label as $y_v$, where $y_v=1$ corresponds to the fraudulent label. The noise associated with hidden fraud is asymmetric, i.e., only if $y_v=1$ then $\tilde{y}_v$ might be different from $y_v$. When there is a difference ($y_v=1$ and $\tilde{y_v}=0$), we say $v$ is a hidden fraud case. Our task is to train our base model to be robust given training data with possible hidden frauds.\\


\noindent\textbf{Learning Bayes Optimal Distribution.} A Bayes optimal label is defined as $y^*_v=\arg\max_y P(y|x_v), (x_v,y)\sim \mathcal{D}$. The Bayes optimal label can be seen as the prediction label of an optimal classifier learned from the clean distribution $\mathcal{D}$. The distribution of $(x,y^*)$ is the Bayes optimal distribution, denoted by $ \mathcal{D}^*$. Based on these concepts, \cite{yang2022estimating} proposes that the label noise problem can be mitigated if we are able to infer the Bayes optimal distribution statistically.

The center of such a Bayes optimal distribution  inference process \cite{yang2022estimating} is the \textit{Bayes-label transition matrix}, which denotes probabilities $P(\tilde{y_v}|y_v^*,x_v) $ that Bayes optimal labels flip to noisy labels for a given instance. A \textit{Bayes label transition model} can be trained to generate the transition matrix associated with each $x_v$. Once we know the Bayes-label transition matrix of training instances, forward \textit{loss correction} \cite{patrini2017making} can be used to train a label-noise robust prediction model. We develop our training method based on this framework, taking into account the characteristics of our problem as follows:

\begin{itemize}
    \item Asymmetric Noise Structure: The Bayes-label transition matrix should  have $P(\tilde{y_v}=1|y_v^*=0,x_v)=0$ and $P(\tilde{y_v}=0|y_v^*=0,x_v)=1$. Intuitively, these two equations imply that if a company does not commit fraud, it will never be reported as fraudulent by regulators. As a result, we only need to model the probability $\gamma_v=P(\tilde{y_v}=0|y_v^*=1,x_v)$ in the Bayes-label transition matrix. We refer to $\gamma_v$ as the hidden fraud rate associated with node $v$.
    \item Instance and Neighborhood Dependent Noise (IND): Hidden fraud is not only instance-dependent noise  but also neighbor-dependent as a fraudulent company may be influenced by the neighbor company. In other words, the Bayes label transition model should model $P(\tilde{y_v}|y_v^*,x_v,\mathcal{N}_v)$, i.e. it depends on the neighbor $N_v$ (on FKG) besides $x_v$. 
\end{itemize}



\subsection{Training Bayes Label Transition Model} \label{sec:stageI}

\subsubsection{Collecting Bayes Optimal Labels} To train the transition model, we need to estimate Bayes optimal labels for company instances in our dataset. The  procedure in  \cite{zhu2021second,cheng2020learning} is adopted to collect such labels as it has been proven effective with non-iid data likes ours. The main idea is to define a confidence regularizer to train a \textit{reference model}, which is used to assign estimated optimal labels to company instances using confidence scores. During this process, some instances can be dropped if the reference model is not confident enough. Here, we train our base model with cross-entropy loss and confidence regularizer \cite{zhu2021second,cheng2020learning} to obtain the reference model. Note that this model is not used in the later stage, only for collecting the Bayes optimal labels. Please refer to \cite{zhu2021second,cheng2020learning} for more details. 


\subsubsection{Training Bayes Label Transition Model}
 After collecting the Bayes optimal labels, we attain a set of filtered examples as $\{(x_i,\tilde{y},\hat{y}^*_i)\}$ where $\hat{y}^*_i$ denote the estimated Bayes optimal labels. Although we can directly use these samples for training a detection model that is robust to some extent \cite{cheng2020learning,cheng2020learning2}, making use of the Bayes-label transition matrix helps us better bridge the Bayes optimal distribution and the noise distribution, resulting in a more robust model \cite{yang2022estimating}.
 
 To better model hidden fraud, which is instance and neighbor dependent noise, we leverage GCN as our transition model due to its ability to incorporate neighbor information for prediction. Specifically, we use a single layer of MW-GCN to estimate $\gamma_v$,
$$
\hat{\gamma_v}=\hat{P}(\tilde{y_i}=0|y_i^*,X^{att}_v,\mathcal{N}_i) =\text{MW-GCN}(X^{att},\mathcal{G}^{sum};\theta)
$$
where $\theta$ is the parameters of MW-GCN. For simplicity,  we do not introduce knowledge embedding and attention fusion at this stage, instead, we use MW-GCN on the sum-up graph $\mathcal{G}^{sum}$, which is obtained by summing all meta-path sub-graphs adjacency matrices. \\

\noindent\textbf{Training.} we use the following  empirical risk to train our transition model on $\{(x_i,\tilde{y},\hat{y}^*_i)\}$:

$$
\mathcal{L}_{IDN}=\frac{1}{m}\sum_{i=1}^m\tilde{y}_i\log(\hat{y}^*_i*\hat{\gamma_i}) *\mathbb{I}(\tilde{y_i}=0)
$$
where $\hat{\gamma}_v$ is the predicted hidden fraud rate associated with node $v$, $\mathbb{I}$ is the indicator function, and $m$ is the number of all filtered examples. Here, different from \cite{yang2022estimating}, $\mathbb{I}$ is used to introduce the asymmetric noise structure into learning the transition model.  After training, we can use the transition model to predict $\hat{\gamma_i}$ for all instances in the training dataset. Note that, the transition model is fixed, i.e., it is not trained along with the fraud detection model in the next stage.


\subsection{Training Hidden Fraud Robust Detection Model} \label{sec:stageII}
Given the estimated hidden fraud rate $\hat{\gamma}_v$ predicted by the Bayes Label Transition Model trained in the previous subsection, we can get the estimated transition matrix $\hat{T}^*(v)$ for each $v$ in the training set $\mathcal{V}^c_t$:
$$\hat{T}^*(v)=
\left(
\begin{array}{cc}
   1  & 0 \\
    \hat{\gamma}_v  & 1-\hat{\gamma}_v
\end{array}
\right)
$$
 where each cell $(i,j)$ corresponds to the transition probability $P(\tilde{y_v}=i|y_v^*=j)$. The transition matrices are then fixed and used to perform forward loss correction \cite{patrini2017making} to train our model \modelname{} as follows:
$$
\mathcal{L}=-\frac{1}{\left|\mathcal{V}^{c}_{t}\right|}\sum_{v \in\mathcal{V}^{c}_{t}} \mathbf{\tilde{y}}_v \log (\mathbf{\hat{y}}_v\cdot\hat{T}(v)^*)
$$
where $\mathbf{\hat{y}}_v \in \mathcal{R}^{1\times2}$ is the predicted output of our base model (see Section \ref{sec:base-method}), $\mathbf{\tilde{y}}_v$ is the one-hot vector encoding of the noisy label $\tilde{y}_v$, $\left|\mathcal{V}^{c}_{t}\right|$ stands for the number of training examples. Note that such loss correction will only take effect when $\tilde{y}_v=1$ due to the special form of our Bayes-label transition matrix. 

\section{Experiments}
This section examines the performance of \modelname{} on our collected datasets (MBM, SME, GEM). Our experiments are designed to answer the following questions:
\begin{itemize}
    \item \textbf{Q1}: Does \modelname{} outperform other possible approaches on the real-world corporate fraud detection datasets? 
    \item \textbf{Q2}: Is the knowledge embedding in \modelname{} essential in dealing with the information overload problem caused by the excessive number of support nodes? 
    \item \textbf{Q3}: Is the hidden fraud robust learning in \modelname{} effective in dealing with the hidden fraud issue compared to other label-noise robust training methods? 
\end{itemize}

\noindent\textbf{General Experimental Settings.} We exploit \textbf{AUC} as the evaluation metric following the common practice in fraud detection studies \cite{bi2022company,li2021live}. We randomly split each of our datasets into train/valid/test following the proportion of \textbf{6:2:2} with the additional constraint that the test set contains no hidden fraud. Specifically, we include only the non-fraud company instances labeled at least 8 years ago in the test set (see Section \ref{sec:data analyse}), while trying to keep train/valid/test in proportion. For data preprocessing, we do min-max normalization on the financial attributes and replace a missing value with the attribute mean. Such a process is common for all the compared methods. We run all experiments 5 times with different random seeds and report the average results. 





\subsection{Comparisons with Baselines (Q1)}\label{sec:q1}
\subsubsection{Experimental Design} Three groups of possible baseline methods are considered as follows (see Appendix for more details): 

\begin{itemize}
    \item \textbf{Traditional methods.} XGBoost \cite{chen2016xgboost} and Deep Neural Network (DNN) are the commonly used traditional methods for fraud detection. These methods use only financial attributes without relational (graph) information.
    \item\textbf{GNN-based baselines.} We include six GNN (Graph Neural Network) baselines in our study - three homogeneous methods namely GCN, MW-GCN, DAGNN \cite{craja2020deep}  and four heterogeneous methods including FastGTN \cite{yun2022graph}, MHGCN \cite{yu2022multiplex}, Tribe-GNN \cite{bi2022company}cand one cost-sensitive baseline CSGNN \cite{csgnn}. Here, MW-GCN is a GCN variant where we introduce Multi-path Weights Matrix into the GNN as described in Section \ref{sec:base-method}. Additionally, all the homogeneous methods were used on the sum-up graph, which is essentially the company subgraphs combined together. Tribe-GNN, on the other hand, considers a hierarchical structure to learn company representations for risk assessment. We adopt the Tribe-GNN model to our study by forming tribes of companies with support nodes and connecting tribes with meta-paths. In all GNN models, we exploit financial attributes for initializing the company nodes.
    \item\textbf{\modelname{} and its variants.} Several variants of our method are considered for comparisons:  (1) ``w/o attn'' means we replace the hierarchical attention fusion with a simple sum; (2) ``w/o KE'' indicates that we use only financial attributes; (3) ``w/o attr'' means we exploit only the knowledge embeddings; (4) ``w/o robust'' means we do not exploit our robust two-stage training, instead just train our model with Cross-Entropy loss on noisy data. 
\end{itemize}

\subsubsection{Experimental Results}
 \begin{table}
\centering
\caption{Main experiment results. We compare our method and its variants with various strong baselines in related problems. The mean (standard error) of AUC is reported.}
\resizebox{\columnwidth}{!}{%
\begin{tabular}{l|c|c|c}
\toprule
                                   &     \textbf{MBM}       & \textbf{SME} & \textbf{GEM} \\ \midrule
XGboost                            & 0.706 (0.003)                 & 0.829 (0.012)                    & 0.766 (0.024)                    \\
DNN                                & 0.665 (0.009)                 & 0.798 (0.034)                    & 0.799 (0.012)                 \\ \midrule
GCN                             & 0.662 (0.018)                  & 0.850 (0.003)                    & 0.805 (0.016)                  \\
MW-GCN                             & 0.674 (0.015)                  & 0.856 (0.007)                    & 0.823 (0.011)                  \\
DAGNN                              & 0.695 (0.009)                  & 0.855(0.004)                    & 0.816 (0.014)                  \\
MHGCN                              & 0.694 (0.017)                  & 0.863 (0.018)                    & 0.800 (0.045)                  \\
Tribe-GNN                             & 0.674 (0.012)                  & 0.868 (0.005)                    & 0.813 (0.007)                  \\
CSGNN                             & 0.646 (0.014)                  & 0.828 (0.012)                    & 0.795 (0.002)                  \\
FastGTN                            & 0.725 (0.007)                  & 0.875 (0.009)                    & 0.845 (0.014)            \\ \midrule
\modelname                         & {\ul 0.782 (0.022)}            & \textbf{0.882 (0.014)}           & \textbf{0.868 (0.015)}   \\
 {\small w/o atten}      & \textbf{0.783 (0.011)}         & 0.875 (0.013)                    & 0.841 (0.019)               \\
  {\small w/o KE  }       & 0.733 (0.028)                  & 0.876 (0.011)                    & 0.828 (0.004)                     \\
  {\small w/o attr }      & 0.728 (0.019)                  & 0.709 (0.035)                    & 0.637 (0.060)                    \\
  {\small w/o robust  }      & 0.743 (0.029)                 & {\ul 0.880 (0.012)}              & {\ul 0.850 (0.009)}           \\ \bottomrule
\end{tabular}%
}

\label{tab:main result}
\end{table}

The average AUC results can be seen from table \ref{tab:main result}. We can draw several conclusions from the main results:\\

\noindent\textbf{Overall.} our method outperforms all baselines and achieves the best AUC across all the datasets, suggesting that our model can better handle the challenges of corporate fraud detection. Particularly, our method outperforms other methods by a large margin in the MBM dataset, which is the one with the most serious hidden fraud problem (see section \ref{sec:data analyse}). Note also that, MBM represents the biggest and the most important market in China A-share market. \\

\noindent\textbf{Regarding Relational Information.} GNN-based methods generally perform better than DNN, showing that relational information between companies can help improve detection results. MW-GCN is better than GCN, suggesting that despite its simplicity, our multi-path weighted matrix is effective for our problem. Overall, \modelname{} outperforms  even the strongest baseline (FastGTN) by a large margin in all three datasets, thanks to its ability the handle the information overload and the hidden fraud problems. Compared to GCN, Tribe-GNN has a higher AUC score, showing that vanilla GNN is not as effective. However, Tribe-GNN's assumption about the graph structure may lead to information loss. Our approach produces better results on three datasets even without two-stage training. \\


\noindent\textbf{Ablation Study.} The ablation study results in Table \ref{tab:main result} reveal several observations. First, both the knowledge embedding and the hidden-fraud robust learning play essential roles in our model as the performance drops without either of them (``w/o KE'' and ``w/o robust''), particularly on the MBM dataset. Second, financial attributes still are the most essential features for corporate fraud detection because the performance drops the most with ``w/o attr''. Third, the attention fusion layer in \modelname{} does not show its advantages on the MBM dataset. It might be because the dataset is big enough to learn good knowledge embeddings, and thus a simple summation can work well for MBM. This is supported by the fact that, in comparison with SME and GEM, the roles of KE and attributes are more balanced on MBM (the performance of ``w/o KE'' is closer to ``w/o attr'' on MBM). However, it is recommended to include the attention fusion layer as it generally performs well on SME and GEM, and 
comparably on MBM. 


\subsection{Impacts of Knowledge Embedding (Q2)}\label{sec:q2}
\begin{figure}
    \centering
    \includegraphics[width=0.95\columnwidth]{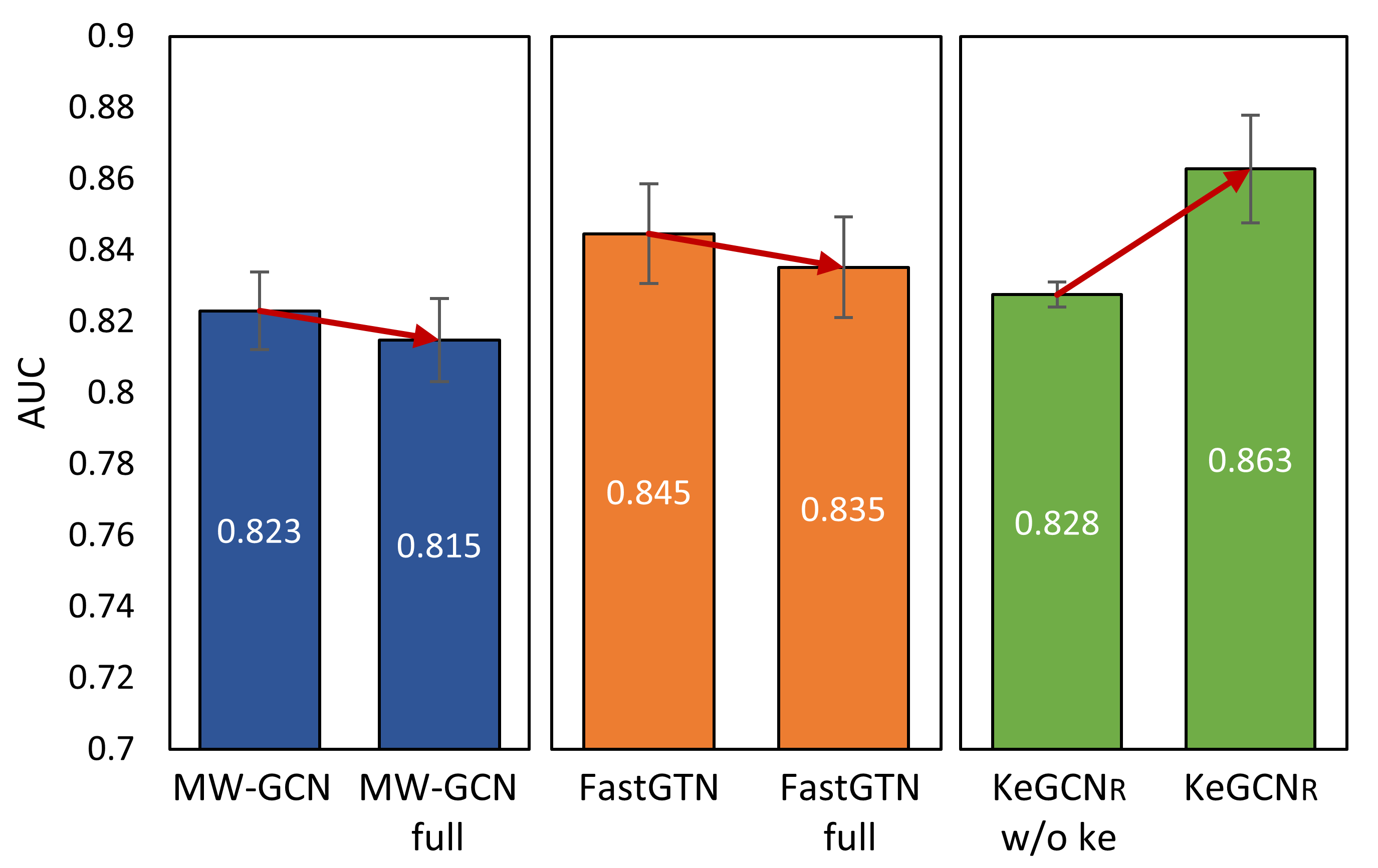}
    \caption{Knowledge embedding (KE) effect on GEM; ``FastGTN", ``MW-GCN", ``\modelname w/o ke" do NOT encode support node information and are run on the company subgraphs; ``FastGTN full", ``MW-GCN full" are run on the full graph with all support nodes; ``\modelname" is the proposed model which uses KE to distill support node information.  }
    \label{fig:kg effect}
\end{figure}

\subsubsection{Experimental Design} This section aims to shed light on the challenge caused by the excessive number of support nodes, the information overload problem. We would like to show that (1) the problem is real, and (2) knowledge embedding pretraining allows us to avoid information overload and better distill knowledge from the support nodes to the company nodes. Towards this goal, we compare \modelname{} and \modelname{} (w/o KE), MW-GCN and FastGTN with a number of additional baselines as follows:  
\begin{itemize}
    \item\textbf{$\text{FastGTN}_\text{full}$.} We apply FastGTN to the \textit{full graph} with all the support nodes (the FKG) of the GEM dataset. FastGTN is chosen because it is the strongest baseline in the main experiment. Attributes of the support nodes are processed by one-hot encoding and then used as node initialization.
    \item\textbf{$\text{MW-GCN}_\text{full}$.} We use MW-GCN as the base method but apply the sampling method like GraphSAGE \cite{hamilton2017inductive}. The resultant model is also tested on the FKG of GEM. Because the model can not process heterogeneous information, we consider that all edges are of the same type, and the support nodes are initialized with zero vectors of the same dimension with company node initial representations.    
\end{itemize}

Note that we picked the GEM dataset because FastGTN can not be applied directly to very large datasets such as the FKGs of the MBM and SME datasets.

\subsubsection{Experimental Results}
 The results are shown in Figure \ref{fig:kg effect}, where two observations can be made. First, the existence of support nodes on the full graph worsens the performance of both MW-GCN and FastGTN. This provides pieces of evidence for our intuition that directly applying GNN on the full graph is not effective due to the information overload problem. Second, \modelname{}, which incorporates information from the neighbor (support) nodes via knowledge embeddings,  is significantly better than \modelname (w/o KE). This suggests that only our method can effectively distill knowledge from support nodes to help improve fraud detection results.

\subsection{Impacts of Robust Training (Q3)} \label{sec:q3}

\subsubsection{Experimental Design} The ablation study has shown that hidden-fraud robust learning is essential for corporate fraud detection, particularly on the MBM dataset where the problem is most serious. This section aims to evaluate our method against other noise-robust learning methods: 
\begin{itemize}
    \item \textbf{Label-noise robust GNN methods.} NRGNN \cite{dai2021nrgnn}, RTGNN \cite{qian2022robust} are the recently proposed methods for handling label noise on graph data with GCN as the base method. We apply them on our sum-up graph like other homogeneous GCN methods.
    \item\textbf{CORES \cite{cheng2020learning}.} We use \modelname{} as the base model and incorporate CORES \cite{cheng2020learning} for loss correction. Specifically, after Bayes optimal label collection, we train \modelname{} with confidence regularization. The model does not use the Bayes label transition model. 
    \item\textbf{BLTM  \cite{yang2022estimating}.} This is the method that uses the Bayes-label transition matrix for loss correction. This method differs from ours in two points: 1) for Bayes optimal label collection, it does not exploit a method that can handle non-idd data; 2) it uses DNN for transition model, which can not handle the neighbor-dependent issue. 
\end{itemize}

For both CORES and BLTM, we also modify the loss function to handle the asymmetric noise structure in our problem. The results of GCN and \modelname{} (w/o robust) are reported for reference.
\begin{table}
\centering
\caption{Comparisons of our robust-learning methods and other alternatives. Those with $\dag$ are base models without label noise robust learning.}
\resizebox{\columnwidth}{!}{
\begin{tabular}{l|l|l|l}
\toprule
                           & \textbf{MBM}           & \textbf{SME}          & \textbf{GEM} \\ \midrule
\multicolumn{1}{l|}{NRGNN}  &  0.688 (0.008)                 &  0.752 (0.011)                    & 0.765 (0.012)                  \\
\multicolumn{1}{l|}{RTGNN}  & 0.696 (0.007)                  &  0.743 (0.008)                   & 0.760 (0.009)       \\           
\multicolumn{1}{l|}{GCN $\dag$}    & 0.662 (0.018)                  & 0.850 (0.003)                    & 0.805 (0.016)                  
\\ \midrule
\multicolumn{1}{l|}{\modelname{}}   & \textbf{0.782 (0.022)}         & \textbf{0.882 (0.014)}           & \textbf{0.868 (0.015)}          \\
\multicolumn{1}{l|}{- w/ BLTM}  & 0.745 (0.069)                  & 0.873 (0.006)                 & 0.847 (0.010)              \\
\multicolumn{1}{l|}{- w/ CORES} & 0.770 (0.005)                  &  0.879 (0.015)               & 0.855 (0.016)      \\
\multicolumn{1}{l|}{- w/o robust$\dag$}  & 0.743 (0.029)                 & 0.880 (0.012)              &  0.850 (0.009)  \\\bottomrule
\end{tabular}%
}

\label{tab:glc effect}
\end{table}

\subsubsection{Experimental Results}
~\
\noindent\textbf{Comparisons with Label-Noise Robust GNN.} A direct comparison with NRGNN and RTGNN is unfair since they are not designed to handle heterogeneous data like \modelname{}, however, we can compare their performance with the base model (GCN). Experiments show that NRGNN and RTGNN, despite being better than GCN on MBM, are inferior to GCN on SME and GEM. The reasons are two folds: 1) the methods are proposed to handle noises in sparsely labeled graphs, which is not the case with our data; 2) they are unaware of the asymmetric noise structure in our data. Without this consideration, a model that corrects hidden fraud could easily fail to handle the class imbalance problem, resulting in a worse AUC.\\

\noindent\textbf{Comparison with other loss correction methods.} Our methods outperform all baselines on three datasets. BLTM is better than the base model \modelname{} (w/o robust) on MBM but fails to do so on SME and GEM. This shows the importance of exploiting a non-idd method for collecting Bayes optimal labels, and explicitly handling IND noise. CORES performs quite well compared to BLTM since it is able to handle non-idd data in collecting Bayes optimal labels. However, CORES fails to perform better than our method, showing that estimating the Bayes-label transition matrix is necessary for bridging the Bayes optimal distribution and the noise distribution. 
\section{Conclusion}

This paper studies the problem of corporate fraud detection. The problem has gained significant attention from regulators and investors in recent years, yet current solutions are still far from sufficient. We collect real-world datasets, which contain a large number of company instances and relations from the Chinese stock markets. Our data analysis reveals two main challenges associated with our data and problem: the information overload issue and the hidden fraud issue. We then propose a novel Knowledge-enhanced GCN model with Robust two-stage learning, \modelname{}, to systematically handle the issues. Numerous experiments show that \modelname{} is better than contemporary graph-based solutions thanks to its ability to handle the information overload issue using knowledge graph embeddings and the hidden fraud issue using our robust training method. In addition, by considering the distinct characteristics associated with hidden fraud, i.e. asymmetric noise structure and Instance and Neighbor Dependence (IND), our robust training solution is shown to perform better than recent methods for handling label noise, including those designed for graph data and instance-dependent noise. 

In the future, more investigations are needed to improve the effectiveness of corporate fraud detection. These include but are not limited to the investigation of hidden fraud robust solutions with class imbalance consideration or the study of effective methods to handle IND noise. Our datasets, which are also the first real-world graph dataset with label noise, will be public to foster future research on these important issues \footnote{our datasets and the source code are available at \href{https://github.com/wangskyGit/KeHGN-R}{https://github.com/wangskyGit/KeHGN-R}}.

\bibliographystyle{elsarticle-num}
\bibliography{tkde}

\begin{thebibliography}{10}
\expandafter\ifx\csname url\endcsname\relax
  \def\url#1{\texttt{#1}}\fi
\expandafter\ifx\csname urlprefix\endcsname\relax\def\urlprefix{URL }\fi
\expandafter\ifx\csname href\endcsname\relax
  \def\href#1#2{#2} \def\path#1{#1}\fi

\bibitem{li2010case}
Y.~Li, The case analysis of the scandal of enron, International Journal of Business and Management 5~(10) (2010) 37.

\bibitem{dechow2011predicting}
P.~M. Dechow, W.~Ge, C.~R. Larson, R.~G. Sloan, Predicting material accounting misstatements, Contemporary Accounting Research 28~(1) (2011) 17--82.

\bibitem{dalnial2014detecting}
H.~Dalnial, A.~Kamaluddin, Z.~M. Sanusi, K.~S. Khairuddin, Detecting fraudulent financial reporting through financial statement analysis, Journal of Advanced Management Science 2~(1) (2014).

\bibitem{hajek2017mining}
P.~Hajek, R.~Henriques, Mining corporate annual reports for intelligent detection of financial statement fraud--a comparative study of machine learning methods, Knowledge-Based Systems 128 (2017) 139--152.

\bibitem{song2014application}
X.-P. Song, Z.-H. Hu, J.-G. Du, Z.-H. Sheng, Application of machine learning methods to risk assessment of financial statement fraud: evidence from china, Journal of Forecasting 33~(8) (2014) 611--626.

\bibitem{Hatice2004}
H.~Uzun, S.~H. Szewczyk, R.~Varma, Board composition and corporate fraud, Financial Analysts Journal 60~(3) (2004) 33--43.

\bibitem{KHANNA2015}
V.~Khanna, E.~H. Kim, Y.~Lu, {CEO} connectedness and corporate fraud, The Journal of Finance 70~(3) (2015) 1203--1252.

\bibitem{Jian2010}
M.~Jian, T.~Wong, Propping through related party transactions, Review of Accounting Studies 15 (2010) 70--105.

\bibitem{CLAESSENS20061}
S.~Claessens, J.~P. Fan, L.~H. Lang, The benefits and costs of group affiliation: evidence from east asia, Emerging Markets Review 7~(1) (2006) 1--26.

\bibitem{kipf2016semi}
T.~N. Kipf, M.~Welling, Semi-supervised classification with graph convolutional networks, arXiv preprint arXiv:1609.02907 (2016).

\bibitem{liu2020towards}
M.~Liu, H.~Gao, S.~Ji, Towards deeper graph neural networks, in: Proceedings of the ACM SIGKDD International Conference on Knowledge Discovery \& Data Mining, 2020, pp. 338--348.

\bibitem{yun2022graph}
S.~Yun, M.~Jeong, S.~Yoo, S.~Lee, S.~Y. Sean, R.~Kim, J.~Kang, H.~J. Kim, Graph transformer networks: Learning meta-path graphs to improve gnns, Neural Networks (2022).

\bibitem{yu2022multiplex}
P.~Yu, C.~Fu, Y.~Yu, C.~Huang, Z.~Zhao, J.~Dong, Multiplex heterogeneous graph convolutional network, in: Proceedings of the ACM SIGKDD International Conference on Knowledge Discovery \& Data Mining, 2022, pp. 2377--2387.

\bibitem{hu2019cash}
B.~Hu, Z.~Zhang, C.~Shi, J.~Zhou, X.~Li, Y.~Qi, Cash-out user detection based on attributed heterogeneous information network with a hierarchical attention mechanism, in: Proceedings of the AAAI Conference on Artificial Intelligence, Vol.~33, 2019, pp. 946--953.

\bibitem{ji2022detecting}
Y.~Ji, Z.~Zhang, X.~Tang, J.~Shen, X.~Zhang, G.~Yang, Detecting cash-out users via dense subgraphs, in: Proceedings of the ACM SIGKDD Conference on Knowledge Discovery and Data Mining, 2022, pp. 687--697.

\bibitem{xu2021towards}
B.~Xu, H.~Shen, B.~Sun, R.~An, Q.~Cao, X.~Cheng, Towards consumer loan fraud detection: graph neural networks with role-constrained conditional random field, in: Proceedings of the AAAI Conference on Artificial Intelligence, Vol.~35, 2021, pp. 4537--4545.

\bibitem{li2021live}
Z.~Li, H.~Wang, P.~Zhang, P.~Hui, J.~Huang, J.~Liao, J.~Zhang, J.~Bu, Live-streaming fraud detection: A heterogeneous graph neural network approach, in: Proceedings of the ACM SIGKDD International Conference on Knowledge Discovery \& Data Mining, 2021, pp. 3670--3678.

\bibitem{bi2022company}
W.~Bi, B.~Xu, X.~Sun, Z.~Wang, H.~Shen, X.~Cheng, Company-as-tribe: company financial risk assessment on tribe-style graph with hierarchical graph neural networks, in: Proceedings of the ACM SIGKDD International Conference on Knowledge Discovery \& Data Mining, 2022, pp. 2712--2720.

\bibitem{qian2022robust}
S.~Qian, H.~Ying, R.~Hu, J.~Zhou, J.~Chen, D.~Z. Chen, J.~Wu, Robust training of graph neural networks via noise governance, WSDM (2023).

\bibitem{dai2021nrgnn}
E.~Dai, C.~Aggarwal, S.~Wang., Nrgnn: Learning a label noise resistant graph neural network on sparsely and noisily labeled graphs., in: Proceedings of the ACM SIGKDD International Conference on Knowledge Discovery \& Data Mining, 2021, pp. 338--348.

\bibitem{bordes2013translating}
A.~Bordes, N.~Usunier, A.~Garcia-Duran, J.~Weston, O.~Yakhnenko, Translating embeddings for modeling multi-relational data, Advances in Neural Information Processing Systems 26 (2013).

\bibitem{sun2018rotate}
Z.~Sun, Z.-H. Deng, J.-Y. Nie, J.~Tang, Rotate: knowledge graph embedding by relational rotation in complex space, in: International Conference on Learning Representations, 2018, pp. 338--348.

\bibitem{DGL-KE}
D.~Zheng, X.~Song, C.~Ma, Z.~Tan, Z.~Ye, J.~Dong, H.~Xiong, Z.~Zhang, G.~Karypis, Dgl-ke: training knowledge graph embeddings at scale, in: Proceedings of the International ACM SIGIR Conference on Research and Development in Information Retrieval, SIGIR '20, 2020, p. 739–748.

\bibitem{sun2011pathsim}
Y.~Sun, J.~Han, X.~Yan, P.~S. Yu, T.~Wu, Pathsim: meta path-based top-k similarity search in heterogeneous information networks, Proceedings of the VLDB Endowment 4~(11) (2011) 992--1003.

\bibitem{wang2019heterogeneous}
X.~Wang, H.~Ji, C.~Shi, B.~Wang, Y.~Ye, P.~Cui, P.~S. Yu, Heterogeneous graph attention network, in: The World Wide Web Conference, 2019, pp. 2022--2032.

\bibitem{yang2022estimating}
S.~Yang, E.~Yang, B.~Han, Y.~Liu, M.~Xu, G.~Niu, T.~Liu, Estimating instance-dependent bayes-label transition matrix using a deep neural network, in: International Conference on Machine Learning, PMLR, 2022, pp. 25302--25312.

\bibitem{zhu2021second}
Z.~Zhu, T.~Liu, Y.~Liu, A second-order approach to learning with instance-dependent label noise, in: Proceedings of the IEEE Conference on Computer Vision and Pattern Recognition, 2021, pp. 10113--10123.

\bibitem{cheng2020learning}
H.~Cheng, Z.~Zhu, X.~Li, Y.~Gong, X.~Sun, Y.~Liu, Learning with instance-dependent label noise: a sample sieve approach, in: International Conference on Learning Representations, 2020, pp. 338--348.

\bibitem{cheng2020learning2}
J.~Cheng, T.~Liu, K.~Ramamohanarao, D.~Tao, Learning with bounded instance and label-dependent label noise, in: International Conference on Machine Learning, PMLR, 2020, pp. 1789--1799.

\bibitem{bronstein2017geometric}
M.~M. Bronstein, J.~Bruna, Y.~LeCun, A.~Szlam, P.~Vandergheynst, Geometric deep learning: going beyond euclidean data, IEEE Signal Processing Magazine 34~(4) (2017) 18--42.

\bibitem{li2018deeper}
Q.~Li, Z.~Han, X.-M. Wu, Deeper insights into graph convolutional networks for semi-supervised learning, in: Proceedings of AAAI conference on artificial intelligence, 2018, pp. 338--348.

\bibitem{hamilton2017inductive}
W.~Hamilton, Z.~Ying, J.~Leskovec, Inductive representation learning on large graphs, Advances in neural information processing systems 30 (2017).

\bibitem{craja2020deep}
P.~Craja, A.~Kim, S.~Lessmann, Deep learning for detecting financial statement fraud, Decision Support Systems 139 (2020) 113421.

\bibitem{grn}
Z.~Huang, K.~Li, Y.~Jiang, Z.~Jia, L.~Lv, Y.~Ma, Graph relearn network: Reducing performance variance and improving prediction accuracy of graph neural networks, Knowledge-Based Systems 301 (2024) 112311.

\bibitem{gscp}
M.~Adjeisah, X.~Zhu, H.~Xu, T.~A. Ayall, Graph contrastive multi-view learning: A pre-training framework for graph classification, Knowledge-Based Systems (2024) 112112.

\bibitem{yun2019graph}
S.~Yun, M.~Jeong, R.~Kim, J.~Kang, H.~J. Kim, Graph transformer networks, Advances in Neural Information Processing Systems 32 (2019).

\bibitem{schlichtkrull2018modeling}
M.~Schlichtkrull, T.~N. Kipf, P.~Bloem, R.~v.~d. Berg, I.~Titov, M.~Welling, Modeling relational data with graph convolutional networks, in: European Semantic Web Conference, 2018, pp. 593--607.

\bibitem{shang2019end}
C.~Shang, Y.~Tang, J.~Huang, J.~Bi, X.~He, B.~Zhou, End-to-end structure-aware convolutional networks for knowledge base completion, in: Proceedings of the AAAI Conference on Artificial Intelligence, Vol.~33, 2019, pp. 3060--3067.

\bibitem{yu2021knowledge}
D.~Yu, Y.~Yang, R.~Zhang, Y.~Wu, Knowledge embedding based graph convolutional network, in: Proceedings of the Web Conference, 2021, pp. 1619--1628.

\bibitem{zhang2019heterogeneous}
C.~Zhang, D.~Song, C.~Huang, A.~Swami, N.~V. Chawla, Heterogeneous graph neural network, in: Proceedings of the ACM SIGKDD International Conference on Knowledge Discovery \& Data Mining, 2019, pp. 793--803.

\bibitem{hu2020heterogeneous}
Z.~Hu, Y.~Dong, K.~Wang, Y.~Sun, Heterogeneous graph transformer, in: Proceedings of The Web Conference 2020, 2020, pp. 2704--2710.

\bibitem{liu2021intention}
C.~Liu, L.~Sun, X.~Ao, J.~Feng, Q.~He, H.~Yang, Intention-aware heterogeneous graph attention networks for fraud transactions detection, in: Proceedings of the ACM SIGKDD International Conference on Knowledge Discovery \& Data Mining, 2021, pp. 3280--3288.

\bibitem{beneish1999detection}
M.~D. Beneish, The detection of earnings manipulation, Financial Analysts Journal 55~(5) (1999) 24--36.

\bibitem{mao2022financial}
X.~Mao, H.~Sun, X.~Zhu, J.~Li, Financial fraud detection using the related-party transaction knowledge graph, Procedia Computer Science 199 (2022) 733--740.

\bibitem{motie2024financial}
S.~Motie, B.~Raahemi, Financial fraud detection using graph neural networks: A systematic review, Expert Systems with Applications 240 (2024) 122156.

\bibitem{mienye2024deep}
I.~D. Mienye, N.~Jere, Deep learning for credit card fraud detection: A review of algorithms, challenges, and solutions, IEEE Access (2024).

\bibitem{liu2022user}
C.~Liu, Y.~Gao, L.~Sun, J.~Feng, H.~Yang, X.~Ao, User behavior pre-training for online fraud detection, in: Proceedings of the ACM SIGKDD International Conference on Knowledge Discovery \& Data Mining, 2022, pp. 3357--3365.

\bibitem{natarajan2013learning}
N.~Natarajan, I.~S. Dhillon, P.~K. Ravikumar, A.~Tewari, Learning with noisy labels, Advances in Neural Information Processing Systems 26 (2013).

\bibitem{manwani2013noise}
N.~Manwani, P.~Sastry, Noise tolerance under risk minimization, IEEE transactions on Cybernetics 43~(3) (2013) 1146--1151.

\bibitem{liu2015classification}
T.~Liu, D.~Tao, Classification with noisy labels by importance reweighting, IEEE Transactions on Pattern Analysis and Machine Intelligence 38~(3) (2015) 447--461.

\bibitem{patrini2017making}
G.~Patrini, A.~Rozza, A.~Krishna~Menon, R.~Nock, L.~Qu, Making deep neural networks robust to label noise: A loss correction approach, in: Proceedings of the IEEE Conference on Computer Vision and Pattern Recognition, 2017, pp. 1944--1952.

\bibitem{yao2020dual}
Y.~Yao, T.~Liu, B.~Han, M.~Gong, J.~Deng, G.~Niu, M.~Sugiyama, Dual t: Reducing estimation error for transition matrix in label-noise learning, Advances in Neural Information Processing Systems 33 (2020) 7260--7271.

\bibitem{hao2022model}
S.~Hao, P.~Li, R.~Wu, X.~Chu, A model-agnostic approach for learning with noisy labels of arbitrary distributions, in: International Conference on Data Engineering, IEEE, 2022, pp. 1219--1231.

\bibitem{rossi2021knowledge}
A.~Rossi, D.~Barbosa, D.~Firmani, A.~Matinata, P.~Merialdo, Knowledge graph embedding for link prediction: a comparative analysis, ACM Transactions on Knowledge Discovery from Data 15~(2) (2021) 1--49.

\bibitem{lee2004information}
B.-S. Lee, O.~M. Rui, S.~S. Wang, Information transmission between the nasdaq and asian second board markets, Journal of Banking \& Finance 28~(7) (2004) 1637--1670.

\bibitem{zhu2019comparative}
J.~Zhu, Y.~Wang, C.~Wang, A comparative study of the effects of different factors on firm technological innovation performance in different high-tech industries, Chinese Management Studies 13~(1) (2019) 2--25.

\bibitem{xiao2015learning}
T.~Xiao, T.~Xia, Y.~Yang, C.~Huang, X.~Wang, Learning from massive noisy labeled data for image classification, in: Proceedings of the IEEE Conference on Computer Vision and Pattern Recognition, 2015, pp. 2691--2699.

\bibitem{yang2015embedding}
B.~Yang, S.~W.-t. Yih, X.~He, J.~Gao, L.~Deng, Embedding entities and relations for learning and inference in knowledge bases, in: International Conference on Learning Representations, 2015, pp. 338--348.

\bibitem{trouillon2016complex}
T.~Trouillon, J.~Welbl, S.~Riedel, {\'E}.~Gaussier, G.~Bouchard, Complex embeddings for simple link prediction, in: International Conference on Machine Learning, 2016, pp. 2071--2080.

\bibitem{hadsell2006dimensionality}
R.~Hadsell, S.~Chopra, Y.~LeCun, Dimensionality reduction by learning an invariant mapping, in: Proceedings of the IEEE Conference on Computer Vision and Pattern Recognition, Vol.~2, 2006, pp. 1735--1742.

\bibitem{velickovic2017graph}
P.~Veli{\v{c}}kovi{\'c}, G.~Cucurull, A.~Casanova, A.~Romero, P.~Li{\`o}, Y.~Bengio, Graph attention networks, in: International Conference on Learning Representations, 2018, pp. 338--348.

\bibitem{chen2016xgboost}
T.~Chen, C.~Guestrin, Xgboost: a scalable tree boosting system, in: Proceedings of the ACM SIGKDD International Conference on Knowledge Discovery \& Data Mining, 2016, pp. 785--794.

\bibitem{csgnn}
X.~Hu, H.~Chen, H.~Chen, S.~Liu, X.~Li, S.~Zhang, Y.~Wang, X.~Xue, Cost-sensitive gnn-based imbalanced learning for mobile social network fraud detection, IEEE Transactions on Computational Social Systems (2023).

\end{thebibliography}

\section*{Acknowledgement}
This paper is supported by the National Key Research and Development Program of China (2022AAA010201). It is also supported by the National Natural Science Foundation of China (72071103), namely “The Underlying Mechanism and Dynamic Prediction of the Important Nodes and Linkages for Risk Transmitting in the RPT-networks of Capital Business Group”. It is also supported by the Fundamental Research Funds for the Central Universities (0118/14370107).

\appendix
\section{Detailed experiment settings}
\subsection{Experimental Settings for Section \ref{sec:q1}} \label{appendix:main-exp}
For all compared methods, the following settings remain the same for all of them:
\begin{table}[htbp]
\centering
\begin{tabular}{|c|c|}
\hline
Trian/valid/test split & 6:2:2                  \\ \hline
loss function          & weighted cross-entropy \\ \hline
hidden layer number    & 2                      \\ \hline
hidden units number    & 1000                   \\ \hline
initial embedding      & financial attributes   \\ \hline
\end{tabular}
\caption{Common setting for all compared methods to ensure fairness}
\label{tab:setting}
\end{table}

For the homogeneous GNN model, we also leverage the sum-up company graph, which is the sum-up of all adjacency matrices of different company sub-graphs, as the input of them. For heterogeneous GNN methods, we also leverage the same meta-paths as their input. And detail setting for different baselines are as follows:
\begin{itemize}
    \item \textbf{XGBoost} \cite{chen2016xgboost}: XGBoost is an algorithm based on GBDT (Gradient Boosted Decision Tree), and one of the most successful methods in many data mining competitions. Here, we used the xgboost package\footnote{https://xgboost.readthedocs.io/en/stable/}, where lambda and gamma are set to 10 and 0.1.
    \item \textbf{DNN}: For node classification with GCN, neural network (NN) is often used as the prediction model on top of convolution layers. To study the effectiveness of convolution layers, i.e. the role of structural information, we include a NN with 2 fully connected layers as one of our baselines. We set the learning rate as 0.0001, the activation function as the sigmoid function and the early-stopping epoch as 300.
    \item \textbf{GCN}: GCN is the common baseline for GNN-based works, and we set the learning rate as 0.001 and the early-stopping epoch as 300 for it. All the settings are exactly the same with MW-GCN to ensure fairness.
    \item \textbf{MW-GCN}: MW-GCN is our proposed method with simple modifications to the vanilla GCN. It's also an important component in our proposed methods, so we leverage it as one of our baselines. We set the learning rate as 0.001 and the early-stopping epoch as 300.
    \item \textbf{DAGNN}\cite{liu2020towards}: DAGNN is a strong homogeneous GNN method that learns node representations by adaptively incorporating information from large receptive fields. It's the recent SOTA GNN methods so we use it as the baseline to better show our model's superiority. We adopt the code in \href{https://github.com/divelab/DeeperGNN}{github}. The final parameters after tuning are K=2, learning rate=0.0001 and early-stopping epoch as 300.
    \item \textbf{MHGCN}\cite{yu2022multiplex}: MHGCN is a recent heterogeneous graph convolutional network that effectively integrates both multi-relation structural signals and attributes semantics into the learned node embeddings. We adopt the code in \href{https://github.com/NSSSJSS/MHGCN}{github}. The final parameters after tuning are learning rate=0.001, epochs=500, degree=2, and per=1.
    \item \textbf{FastGTN} \cite{yun2022graph}: FastGTN is the improved version of GTN and has the ability to find new meta-paths. It achieved SOTA results on both homogeneous and heterogeneous graph benchmarks. We adopt the code in \href{https://github.com/seongjunyun/Graph_Transformer_Networks}{github}. The final parameters after tuning are learning rate=0.0001, epochs=2000, channels number=2, K=1 and early-stopping epoch=300.
    \item \textbf{CSGNN} \cite{csgnn}: CSGNN introduces a cost-sensitive graph neural network that employs reinforcement learning to adaptively determine an optimal sampling threshold. By conducting neighbor sampling based on node similarity, the approach effectively mitigates graph imbalance challenges. We leverage code in official \href{https://github.com/xxhu94/CSGNN}{github}. The final parameters after tuning are learning rate=0.002, $\lambda$ in loss function is 0.2 and early-stopping epoch as 100.
\end{itemize}

\paragraph{Implementation Details} For {\modelname}, knowledge graph pretraining was conducted by using \cite{DGL-KE}, where we set the embedding dimension of 500, the learning rate of 0.25, and the maximum number of epochs of 80000. For the Bayes transition model learning stage, we set the hidden unit number as 500 for the MW-GCN layer during the hidden-fraud probability learning. For the final end-to-end node classification learning stage with loss correction, we use MW-GCN with $L=2$  layers, the ReLu activation function, and the hidden layer sizes of $N_l=1000$. As an imbalance classification problem, we apply weighted cross-entropy loss for our methods. We tuned hyper-parameters and do early stopping according to the result on the validation set. The source code for \modelname{} can be found in GitHub\footnote{https://github.com/wangskyGit/KeHGN-R}.

For other baselines, we followed the standard practice to tune hyper-parameters on the validation set. Note that, during training, we also apply weighted cross entropy loss to handle the class imbalance issue.

\subsection{Experimental Settings for Section \ref{sec:q2}} \label{appendix:ke-experiment}
For the experiment in section 6.2, the common settings are the same as in table \ref{tab:setting}. The implementing and setting details are as follows:
\begin{itemize}
    \item $\text{FastGTN}_\text{full}$: Categorical support node attributes are processed by one-hot encoding and then used as the initial feature of support nodes. Codes of FastGTN cannot handle nodes with attributes in different feature spaces and we add another linear layer before the FastGTN forward function to make them into the same vector space.  We keep all other settings the same as FastGTN in section 6.1 for a fair comparison.  
    \item $\text{MW-GCN}_\text{full}$: We apply GraphSAGE as the sampling layer with MW-GCN as the aggregation method to the \textit{full graph}. Because the model can not process heterogeneous information, we initialize support node attributes as zero vectors in the same feature space as financial attributes and consider all edge types the same. We leverage graphsage code in \href{https://github.com/twjiang/graphSAGE-pytorch}{https://github.com/twjiang/graphSAGE-pytorch} and remain the model setting of MW-GCN as the same in section 6.1 for a fair comparison. 
\end{itemize}
\subsection{Experimental Settings for Section \ref{sec:q3}} \label{appendix:robust-experiment}
We replace our two-stage hidden fraud robust training (HFRT) with two related label noise-robust methods while remaining the backbone node classification method the same. And we also compare with two other noise-resistant GNNs. The detailed experimental settings of them are as follows:
\begin{itemize}
    \item  NRGNN \cite{dai2021nrgnn}: NRGNN is a novel method of learning noise-resistant GNNs on graphs with noisy and limited labels; It links the unlabeled nodes with labeled nodes of high feature similarity to bring more clean label information. We adopt the code in \href{https://github.com/EnyanDai/NRGNN}{github}. The final parameters after tuning are: edge\_hidden=50, learning rate=0.001, alpha=0.001, beta=0.001, n\_p=5, n\_n=5.
    \item  RTGNN\cite{qian2022robust}: Similar to NRGNN, RTGNN is also a model for graphs with noisy and scarce labels. It introduces self-reinforcement and consistency regularization as supplemental supervision and achieves SOTA results on simulated noisy datasets. We leverage the code in \href{https://github.com/GhostQ99/RobustTrainingGNN}{github}. The final parameters after tuning are: edge\_hidden=50, learning rate=0.001, alpha=0.1, K=20, n\_neg=5, decay\_w=0.8, co\_lambda=0.1. 
    \item CORES\cite{cheng2020learning}: To ensure fairness, we leverage the same hyper-parameter setting of sample sieve in CORES as the HFRT in our model \modelname. And settings for the backbone fraud detection model are also the same.
    \item BLTM\cite{yang2022estimating}: The raw method uses ResNet as the backbone model because they run experiments on computer vision benchmarks. We leverage \backboneModel{} as the backbone model and change the Bayes-layer transition matrix learning model to a two-layer DNN with 500 hidden units and the sigmoid activation function. And settings for the backbone fraud detection model are the same for fairness.
\end{itemize}

\end{document}